\documentclass[letterpaper]{article} 
\usepackage{aaai2026}  
\usepackage{amsmath} 
\usepackage{times}  
\usepackage{helvet}  
\usepackage{courier}  
\usepackage[hyphens]{url}  
\usepackage{graphicx} 
\usepackage{natbib}  
\usepackage{caption} 
\usepackage{booktabs} 
\usepackage{algorithm}
\usepackage{algorithmic}
\usepackage{multirow}
\usepackage{newfloat}
\usepackage{listings}
\DeclareCaptionStyle{ruled}{labelfont=normalfont,labelsep=colon,strut=off} 
\floatstyle{ruled}
\newfloat{listing}{tb}{lst}{}
\floatname{listing}{Listing}
\title{AdaFusion: Prompt-Guided Inference with Adaptive Fusion of Pathology Foundation Models}
\newcommand{\corresponding}{\textsuperscript{$\ddagger$}}
\author{
    Yuxiang Xiao\textsuperscript{\rm 1}\equalcontrib,
    Yang Hu\textsuperscript{\rm 2}\equalcontrib\corresponding,
    Bin Li\textsuperscript{\rm 3},
    Tianyang Zhang\textsuperscript{\rm 3},
    Zexi Li\textsuperscript{\rm 4},
    Huazhu Fu\textsuperscript{\rm 5},
    Jens Rittscher\textsuperscript{\rm 2} \textsuperscript{\rm 6},
    Kaixiang Yang\textsuperscript{\rm 1}\corresponding,
}

\affiliations{
    \textsuperscript{\rm 1}School of Computer Science and Engineering, South China University of Technology, Guangzhou, China\\
    
    \textsuperscript{\rm 2}Nuffield Department of Medicine, University of Oxford, Oxford, UK\\

    \textsuperscript{\rm 3}Department of Engineering Science, University of Oxford, Oxford, UK\\

    \textsuperscript{\rm 4}St Catherine's College, University of Oxford, Oxford, UK\\
    
    \textsuperscript{\rm 5}Institute of High Performance Computing, Agency for Science, Technology and Research,  Singapore\\

    \textsuperscript{\rm 6}Department of Engineering Science, University of Oxford, Oxford, UK\\
}

\usepackage{bibentry}
\newcommand{\first}[1]{\textbf{\underline{#1}}}
\newcommand{\second}[1]{\textbf{#1}}
\newcommand{\third}[1]{\textbf{\textit{#1}}}
\usepackage{booktabs}
\usepackage{multirow}
\usepackage{amsmath}
\usepackage{amssymb}

\begin{document}
\begin{sloppypar}
\maketitle

\begin{abstract}
Pathology foundation models (PFMs) have demonstrated strong representational capabilities through self-supervised pre-training on large-scale, unannotated histopathology image datasets. However, their diverse yet opaque pretraining contexts, shaped by both data-related and structural/training factors, introduce latent biases that hinder generalisability and transparency in downstream applications.
In this paper, we propose AdaFusion, a novel prompt-guided inference framework that, to our knowledge, is among the very first to dynamically integrate complementary knowledge from multiple PFMs. Our method compresses and aligns tile-level features from diverse models and employs a lightweight attention mechanism to adaptively fuse them based on tissue phenotype context.
We evaluate AdaFusion on three real-world benchmarks spanning treatment response prediction, tumour grading, and spatial gene expression inference. Our approach consistently surpasses individual PFMs across both classification and regression tasks, while offering interpretable insights into each model’s biosemantic specialisation. These results highlight AdaFusion’s ability to bridge heterogeneous PFMs, achieving both enhanced performance and interpretability of model-specific inductive biases.
\end{abstract}


\section{Introduction}
\label{sec1}

High-resolution digital pathology images provide rich tissue context and cellular detail essential for computational modelling in cancer diagnosis, grading, and prognosis~\cite{lee2025benchmarking}. Across a wide spectrum of pathology image analysis tasks, from patch-level phenotype prediction to slide-level weakly supervised classification, the quality of tile-level features is a critical determinant of downstream performance. In slide-level tasks, weakly-supervised multiple instance learning (MIL) has long served as a cornerstone paradigm \cite{ilse2018attention}, offering a scalable solution for analysing gigapixel whole slide images (WSIs). Through a two-stage pipeline involving tiling and aggregation, WSIs are decomposed into thousands of smaller tiles, each capturing local tissue histological semantics \cite{lu2021data}. Tile-level features, when appropriately extracted, serve as powerful representations for reconstructing global slide-level information \cite{chen2024towards}, significantly influencing model performance in downstream tasks such as subtyping, grading, and prognosis. In tile-level tasks, by contrast, models directly operate on patch-level representations to predict molecular or spatial markers, bypassing the need for global aggregation but placing equal if not greater emphasis on the quality and expressiveness of local features.

Recently, the rise of self-supervised learning (SSL)-based pathology foundation models (PFMs) has revolutionised the extraction of tile representations \cite{chen2024towards,xu2024whole,vorontsov2024foundation}. These models, pretrained on massive pathological image cohorts, provide high-quality and transferable tile embeddings that can easily integrate with multimodal information such as textual clinical reports or genomic profiles \cite{lu2024visual,chen2025visual}. The widespread deployment of PFMs has ushered in a new era in computational pathology, shifting the focus from task-specific learning towards generalisable, feature-centric pipelines \cite{wang2024pathology}.

\begin{figure}[t!]
\centering
\includegraphics[scale=0.3]{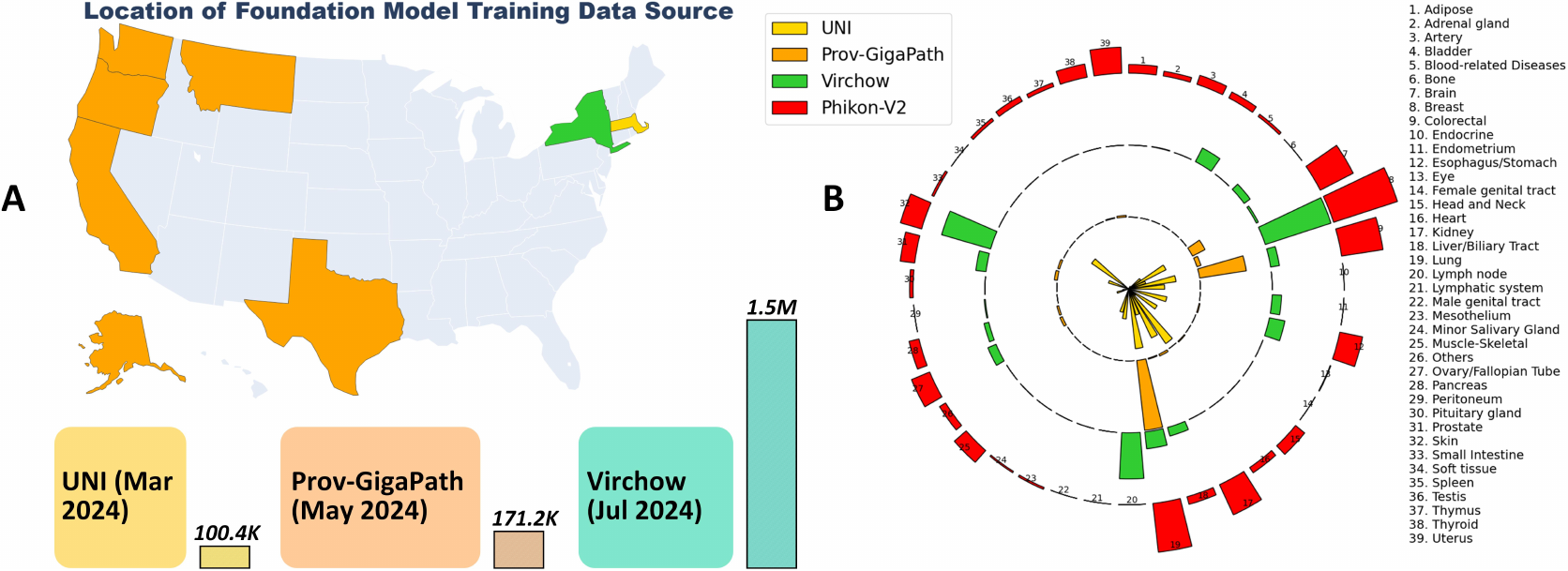}
\caption{Heterogeneous pre-training contexts for pathology foundation models (PFMs). 
A. Geographic sources of training data used by several PFMs.  
B. Imbalanced distribution of cancer types in training datasets across PFMs.}
\label{fig1}
\end{figure}

Despite these advancements, existing PFMs are often developed under diverse and opaque training conditions, introducing potential biases that stem from limited data diversity and inaccessible pretraining sources \cite{chen2024towards,lu2024visual,xu2024whole}. Due to patient privacy regulations, most of the private datasets used to train PFMs are not publicly available, making it difficult to assess their quality, discrepancies, or underlying distributions. While some PFMs claim to be trained on multi-centre datasets, key demographic details, such as sampling geography, ethnicity, and gender, are frequently unknown \cite{vorontsov2024foundation,wang2024pathology}. Even among those with disclosed training sources, the data are typically collected from one or two regional hospitals, and often display pronounced class imbalance across cancer types, following long-tailed distributions \cite{chen2024towards,vorontsov2024foundation}. As illustrated in Figure \ref{fig1}, the growing scale of PFM training data has not fundamentally resolved these issues of domain homogeneity and cancer-type skew.

Beyond data-induced biases, the structural design and training algorithms of PFMs introduce additional representational biases. The widely used Vision Transformer (ViT) processes image patches as independent tokens~\cite{chen2024towards,zimmermann2024virchow2}, promoting local feature learning but limiting long-range spatial modelling. In contrast, emerging dilated ViT architectures~\cite{xu2024whole} emphasise global context and tissue architecture, often at the cost of cytological detail. Methodologically, the choice of self-supervised objective further shapes representations: contrastive approaches like DINOv2 instil discriminative priors, while masked autoencoding (e.g., MAE) encourages holistic, context-aware features. Visual–language models introduce semantic priors by aligning features with language~\cite{lu2024visual}, but may overlook morphology beyond linguistic description. These architectural and training choices yield PFMs with specialised yet inherently limited perspectives on the histopathological landscape.

Such biases are silently propagated through the pretrained feature space, compromising the robustness and fairness of downstream analyses~\cite{du2025ethics}. Given the proliferation of diverse PFMs, integrating their complementary strengths presents an intuitive yet underexplored solution. In this work, we propose a deceptively simple yet efficient and interpretable framework for adaptive feature fusion across multiple PFMs. Our method first compresses the high-dimensional embeddings from each PFM and then concatenates them. To facilitate effective cross-model communication, we introduce a compact inner-attention unit that dynamically modulates feature importance.

We form our adaptive fusion module as \textit{Prompt-Guided Inference}. Without modifying any of the pretrained PFMs, we learn a lightweight attention-driven prompt mechanism that adaptively reweights and integrates features according to tissue phenotype. Our design not only enhances predictive performance but also produces interpretable attention maps that highlight which PFMs contribute most to specific morphological phenotypes.

With validation on multiple computational pathology benchmarks, including treatment response prediction, automated tumour grading, and spatial transcriptomic inference. We demonstrate the superior performance and feature efficiency of the proposed method compared to single-PFM baselines, while also offering rich interpretability into the biosemantic-aimed strengths of different PFMs.
In summary, our contributions are three-fold:

\begin{itemize}
\item We propose a prompt-guided inference framework that adaptively fuses features from multiple PFMs. A lightweight attention module grabs inter-PFMs communication based on tissue phenotype, while compressing high-dimensional embeddings for improved efficiency.
\item We provide interpretable modelling of dynamic feature importance, conditioned on tissue phenotypes. This allows us to visualise the contribution of each PFM to different morphological phenotypes, offering insights into the clinical and biological applicability of PFMs.
\item We conduct extensive multi-perspective validation on various real-world tasks. Our approach demonstrates decent improvements in performance and interpretability.
\end{itemize}

\section{Related Work}
\label{sec2}

\begin{figure*}[ht!]
\centering
\includegraphics[width=0.8\textwidth]{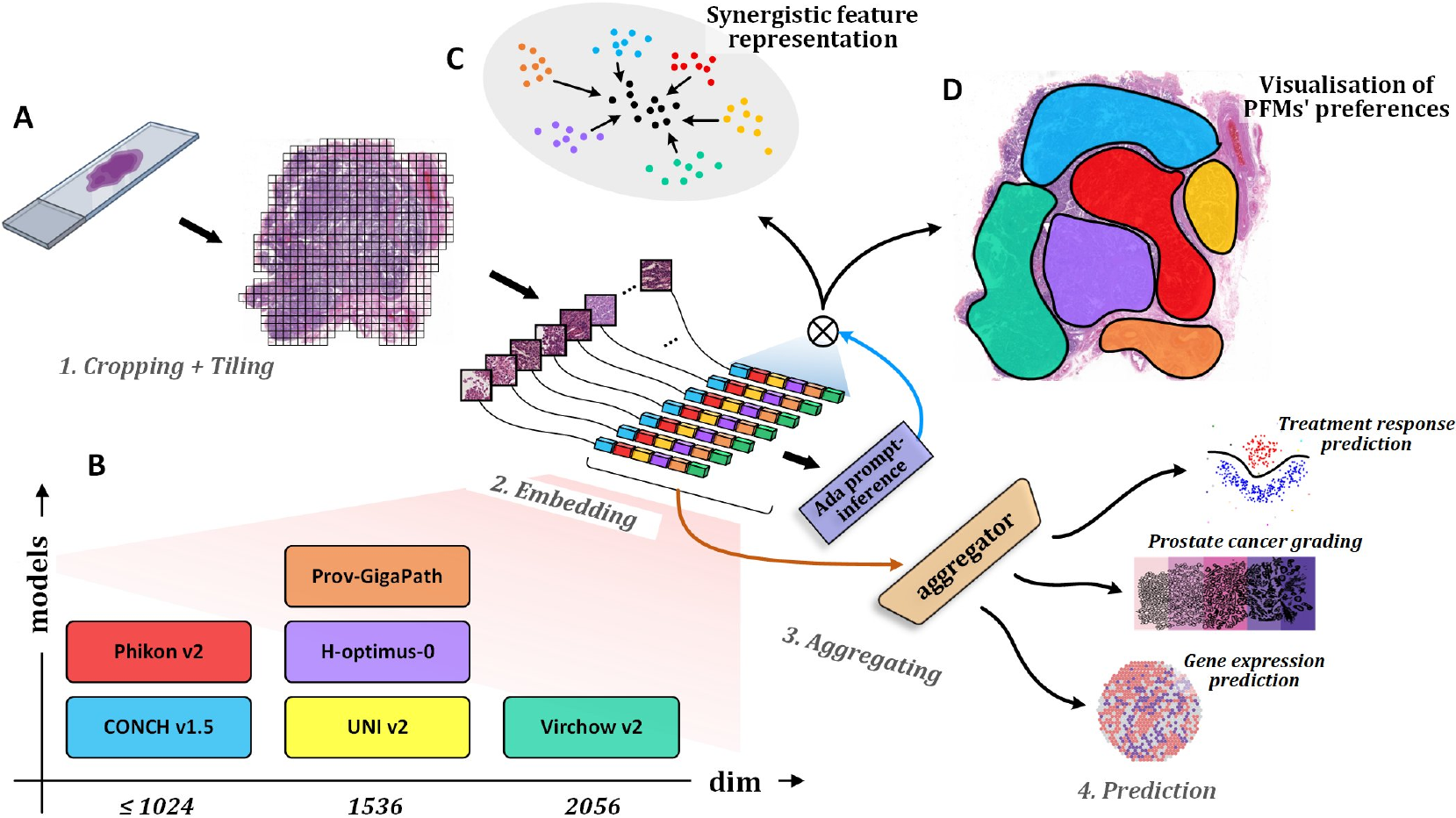}
\caption{
Overview of the prompt-guided inference framework for adaptive fusion of PFMs. 
A. Digital pathology analysis workflow: WSIs are partitioned into tiles, from which features are extracted and passed through an adaptive fusion and prompt inference module. The resulting features are either directly used for downstream predictions or optionally aggregated for slide-level tasks.
B. Tile-level embeddings from multiple PFMs with varying output dimensions are integrated into a unified representation space.
C. The fused features yield a more synergistic view of histological phenotypes.
D. Interpretation module provides highlighting of spatial regions where different PFMs contribute most effectively to specific morphological phenotypes.
}
\label{fig2}
\end{figure*}

Early works in computational pathology (CPath), especially in a weakly supervised learning setting, often rely on feature extractors pretrained on natural image datasets such as ImageNet~\cite{lu2021data}.
With the emergence of self-supervised learning (SSL), domain-specific feature extractors pretrained on pathological datasets using frameworks like MoCo and SimCLR have demonstrated superior performance~\cite{dehaene2020self, li2021dual}.

More recently, SSL-driven pathology foundation models (PFMs), trained on large-scale unlabelled WSI data, have shown strong potential by learning generalisable visual representations. Early PFM research focused on tile-level feature learning from large cohorts of H\&E-stained WSIs. Representative examples include CTransPath \cite{wang2022transformer}, 
along with UNI \cite{chen2024towards}, Phikon-v2 \cite{filiot2024phikon}, and the Virchow series \cite{zimmermann2024virchow2}, which were pre-trained with DINOv2 \cite{oquab2024dinov2}. More recent models toward increasingly large-scale, with parameters exceeding a billion, such as H-optimus-0 \cite{hoptimus0} and Prov-GigaPath \cite{xu2024whole}. Some PFMs have also integrated multimodal data (e.g., paired clinical reports or multi-omics information) to enable cross-modal alignment and improve biological relevance, as demonstrated by CONCH \cite{lu2024visual} and OmiCLIP \cite{chen2025visual}.

The primary challenges in training PFMs stem from the scarcity of diverse, high-quality data and the stringent privacy constraints associated with medical images. Even with large-scale datasets, generalisation is often limited by pronounced batch effects~\cite{vaidya2024demographic}, which arise from differences in tissue collection, fixation, staining, scanner types, and patient demographics across medical centres. These non-pathological variations can hinder meaningful biological signals in the feature space, causing samples to cluster according to acquisition protocols and centre-specific factors rather than disease-relevant patterns such as cancer subtype~\cite{de2025current}.

Therefore, several strategies were proposed. 
\cite{lee2025benchmarking} demonstrated that fine-tuning PFMs on specific downstream tasks can help models focus more on bio-relevant patterns. 
\cite{guo2025focus} leveraged text-based supervision to mitigate site-specific biases. COBRA~\cite{lenz2025unsupervised} improved the robustness of slide-level aggregators by randomly sampling various PFMs while extracting tile-level features from a single WSI. Additionally, \cite{gao2025features} explored the effectiveness of combining features from three distinct PFMs to boost downstream task performance.
However, these approaches do not adaptively fuse features based on tissue phenotype, often ignore the computational burden of combining high-dimensional embeddings, and provide limited interpretability.

\section{Method}
\label{sec3}

\subsection{Overall Framework}
We propose \textbf{AdaFusion}, a prompt-guided fusion framework for integrating features from multiple pre-trained PFMs. These PFMs are treated as frozen prior knowledge sources, while a lightweight Prompt Tuner dynamically recalibrates their features. It produces sample-specific attention weights to modulate each model’s contribution, enabling task-adaptive inference without updating PFM parameters. As shown in Figure~\ref{fig2}, AdaFusion consists of three stages: (1) multi-source prompt extraction and compression, (2) dynamic prompt composition and tuning, and (3) prompt-based prediction and contribution interpretation.

We begin with WSI preprocessing using TRIDENT~\cite{zhang2025accelerating} to segment foreground tissue from background and artefacts. The segmented regions from high-resolution WSIs are divided into non-overlapping tiles of size $224 \times 224$ pixels. Following the MIL paradigm, each WSI is treated as a bag of tiles. Formally, the $n$-th WSI is represented as a bag $\mathcal{X}n = { \mathbf{x}{n,i} }{i=1}^{M_n}$, where each instance $\mathbf{x}{n,i}$ is an individual tile and $M_n$ is the total number of tiles from that WSI.

\subsection{Multi-Source Prompt Extraction and Compression}
Given an input pathology tile $x \in \mathbf{R}^{H \times W \times C}$, we extract features from a set of $N$ diverse, PFMs $\{\mathcal{F}_1, \mathcal{F}_2, \dots, \mathcal{F}_N\}$. PFM $\mathcal{F}_i$, with frozen parameters $\theta_{\mathcal{F}_i}$, acts as a feature extractor that maps the input to a model-specific embedding $\mathbf{f}_i \in \mathbf{R}^{d_i}$. The feature dimension $d_i$ varies across models.
\begin{equation}
    \mathbf{f}_i = \mathcal{F}_i(x; \theta_{\mathcal{F}_i}).
\end{equation}

To enable effective integration and reduce computational / storage cost, we project these heterogeneous embeddings into a unified, joint embedding space. Each raw feature $\mathbf{f}_i$ is compressed via mean pooling to produce a embedding $\mathbf{e}_i$ with lower dimension $d$ (set to 64 in our implementation):
\begin{equation}
    \mathbf{e}_i = \mathbf{Pool}(\mathbf{f}_i) \in \mathbf{R}^{d},
\end{equation}
where $\mathbf{Pool}(\cdot)$ denotes the mean pooling operator. This yields a set of $N$ standardised embeddings $\{\mathbf{e}_1, \dots, \mathbf{e}_N\}$ of a tile, each encapsulating the core representation from a distinct PFM in a unified form.

\begin{figure}[ht!]
\centering
\includegraphics[scale=0.4]{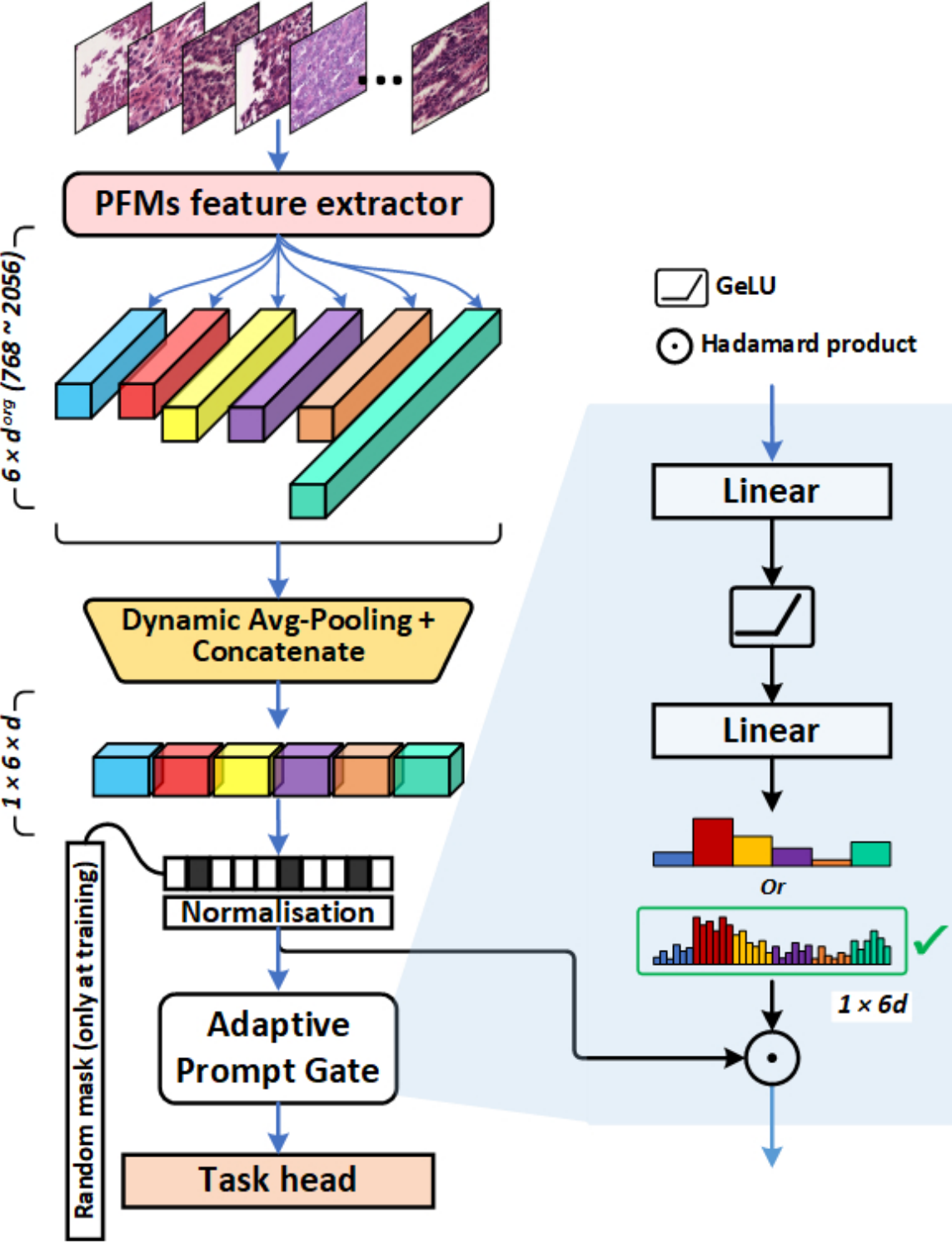}
\caption{Structure of prompt-guided adaptive fusion.}
\label{fig3}
\end{figure}

\subsection{Dynamic Prompt Composition and Tuning}
\label{ssec:tuning}
The standardised embeddings are concatenated to form a compound embedding $\mathbf{E}_{\text{compound}}$ that aggregates multi-source feature knowledge:
\begin{equation}
    \mathbf{E}_{\text{compound}} = [\mathbf{e}_1; \mathbf{e}_2; \dots; \mathbf{e}_N] \in \mathbf{R}^{N \times d}.
\end{equation}
Here, $[\cdot; \cdot]$ denotes the concatenation operator. For $N=6$ and $d=64$, this results in a $384$ - dimensional vector.

To enhance robustness during training, we introduce a random masking strategy on $\mathbf{E}_{\text{compound}}$ by applying an element-wise binary mask $\mathbf{M} \in \{0, 1\}^{N \times d}$, where each entry is retained with probability $1 - \rho$ (we set $\rho = 0.2$). This yields the masked embedding:
\begin{equation}
    \mathbf{E}_{\text{masked}} = \mathbf{E}_{\text{compound}} \odot \mathbf{M}.
\end{equation}
This stochastic regularisation encourages the tuner to avoid overfitting to a subset of feature channels and promotes better utilisation of complementary signals across PFMs.

The core of our fusion mechanism is the Prompt Tuner, a lightweight neural module $\mathcal{T}$ parameterised by $\theta_t$. It takes the masked compound prompt $\mathbf{E}_{\text{masked}}$ as input, learns the interdependencies among multi-source features, and outputs a sample-specific gating matrix $\mathbf{W}_{\text{gate}}$, which is configured differently for our two proposed variants:
\begin{itemize}
    \item For \textbf{AdaFusion-Coarse}, the tuner generates a coarse, source-level gating vector $\mathbf{W}_{\text{gate}}' \in \mathbf{R}^{N \times 1}$, $N$ is the number of PFMs (e.g., $N=6$). Each of the $N$ weights is then broadcast across its corresponding feature block, resulting in the final gating matrix $\mathbf{W}_{\text{gate}} \in \mathbf{R}^{N \times d}$.
    \item For \textbf{AdaFusion-Fine}, the tuner generates a fine-grained, element-wise gating matrix $\mathbf{W}_{\text{gate}} \in \mathbf{R}^{N \times d}$, where $d$ is the full dimension of the concatenated features. This provides a unique weight for every single feature dimension.
\end{itemize}

We implement $\mathcal{T}$ as a two-layer MLP with non-linear activations and normalisation. Regardless of the variant, the resulting gating matrix $\mathbf{W}_{\text{gate}}$ modulates the compound prompt through an element-wise Hadamard product, producing a tuned prompt $\mathbf{P}_{\text{tuned}}$:
\begin{equation}
    \mathbf{P}_{\text{tuned}} = \mathbf{P}_{\text{compound}} \odot \mathbf{W}_{\text{gate}}.
    \label{eq:tuned_prompt}
\end{equation}
This tuned prompt serves as a recalibrated, phenotype-aware fusion of the PFM ensemble, tailored to each input bag $\mathcal{X}_n$.

\subsection{Downstream Prediction and Training Objective}
The final representation $\mathbf{P}_{\text{tuned}}$ is fed into the \textbf{task head} $\mathcal{H}$, which consists of a linear classifier or MIL aggregator with trainable parameters $\theta_h$, to produce the class logits and generate the final prediction $\hat{y}$.
\begin{equation}
    \hat{y} = \mathcal{H}(\mathbf{P}_{\text{tuned}}; \theta_h).
\end{equation}

Overall, the proposed prompt-guided fusion strategy is based on low-dimensional features and introduces only minimal additional parameters, maintains computational efficiency, and is less prone to overfitting.

\subsection{Interpretability via Contribution Analysis}
\label{ssec:interpretability}
A key advantage of our framework lies in its inherent interpretability. The learning gating prompt $\mathbf{W}_{\text{gate}}$ directly reflects the importance from each PFM for a given input. We can quantify the overall contribution of  $\mathcal{F}_i$ by analysing its corresponding segment within the gating prompt.

First, we deconstruct the gating prompt $\mathbf{W}_{\text{gate}}$ back into its constituent sub-vectors $\{\mathbf{w}_1, \dots, \mathbf{w}_N\}$, where each $\mathbf{w}_i \in \mathbf{R}^{d}$ corresponds to the weights applied to the visual prompt $\mathbf{p}_i$ from model $\mathcal{F}_i$. A scalar contribution score $S_i$ is then computed by averaging the values within $\mathbf{w}_i$:
\begin{equation}
    S_i = \frac{1}{d} \sum_{j=1}^{d} (\mathbf{w}_i)_j.
    \label{eq:contribution_score}
\end{equation}
This score provides a concise and interpretable measure of the relevance of $\mathcal{F}_i$ in the final prediction for a specific input. It offers an insightful understanding of how the fusion module dynamically allocates attention across PFMs, revealing the model’s decision-making process in weighting feature sources under different histopathological phenotype contexts.

\section{Results}
\label{sec4}

\begin{table*}[ht!]
\small
\setlength{\tabcolsep}{1mm}
\centering
\begin{tabular}{l|cccc|cccc|cccc|cccc}
\toprule
\multicolumn{1}{c|}{\multirow{2}{*}{\textbf{Model}}} & \multicolumn{8}{c|}{\textbf{ATEC23}} & \multicolumn{8}{c}{\textbf{PANDA}} \\
\cmidrule(lr){2-9} \cmidrule(lr){10-17}
\multicolumn{1}{c|}{} & \multicolumn{4}{c|}{ACC} & \multicolumn{4}{c|}{AUC} & \multicolumn{4}{c|}{ACC} & \multicolumn{4}{c}{AUC} \\
\cmidrule(lr){2-5} \cmidrule(lr){6-9} \cmidrule(lr){10-13} \cmidrule(lr){14-17}
\multicolumn{1}{c|}{} & ori-d & 512 & 256 & 64 & ori-d & 512 & 256 & 64 & ori-d & 512 & 256 & 64 & ori-d & 512 & 256 & 64 \\
\midrule
CONCH v15 & 0.779 & 0.779 & 0.786 & 0.724 & 0.795 & 0.798 & 0.818 & 0.745 & 0.673 & 0.674 & 0.659 & 0.641 & 0.909 & 0.908 & 0.896 & 0.885 \\
Phikon v2 & 0.828 & 0.810 & 0.807 & 0.779 & 0.861 & 0.873 & 0.853 & 0.817 & 0.721 & 0.713 & 0.670 & 0.623 & 0.928 & 0.923 & 0.902 & 0.872 \\
UNI v2 & 0.831 & 0.824 & 0.779 & 0.724 & 0.889 & 0.880 & 0.848 & 0.779 & 0.739 & 0.717 & 0.684 & 0.643 & 0.936 & 0.922 & 0.908 & 0.888 \\
H-optimus-0 & 0.855 & 0.803 & 0.810 & 0.741 & 0.913 & 0.851 & 0.862 & 0.783 & 0.753 & 0.728 & 0.699 & 0.681 & 0.942 & 0.930 & 0.914 & 0.908 \\
Prov-GigaPath & 0.841 & 0.797 & 0.814 & 0.772 & 0.868 & 0.849 & 0.841 & 0.819 & 0.742 & 0.714 & 0.681 & 0.666 & 0.936 & 0.924 & 0.908 & 0.897 \\
Virchow2 & 0.852 & 0.841 & 0.824 & 0.721 & 0.885 & \second{0.905} & 0.878 & 0.776 & 0.740 & 0.722 & 0.692 & 0.657 & 0.938 & 0.928 & 0.911 & 0.892 \\
\midrule
Self-Attn & - & \second{0.855} & \third{0.852} & \third{0.838} & - & 0.890 & \first{0.910} & \third{0.886} & - & 0.727 & \third{0.741} & \third{0.754} & - & 0.926 & \third{0.940} & \third{0.942} \\
Top-3 MoE & - & \third{0.852} & 0.855 & 0.828 & - & 0.873 & 0.896 & 0.880 & - & \third{0.731} & 0.731 & 0.718 & - & \third{0.934} & 0.932 & 0.927 \\
AdaFusion-Coarse & - & 0.841 & \first{0.866} & \first{0.862} & - & \third{0.896} & \third{0.900} & \first{0.901} & - & \second{0.793} & \second{0.793} & \second{0.792} & - & \second{0.952} & \second{0.950} & \second{0.951} \\
AdaFusion-Fine & - & \first{0.869} & \second{0.863} & \first{0.862} & - & \first{0.910} & \second{0.909} & \second{0.900} & - & \first{0.821} & \first{0.817} & \first{0.815} & - & \first{0.958} & \first{0.957} & \first{0.955} \\
\bottomrule
\end{tabular}%
\caption{Performance on classification benchmarks (ATEC23 and PANDA). The table details the performance of individual PFMs with the original dimension features (ori-d) and their features pooled to different dimensions (e.g., 512, 256, 64). It compares them with our AdaFusion variants and other fusion baselines. First, second, and third best results in each column are denoted as bold-underline, bold, and bold-italic.}
\label{tab:classification_results}
\end{table*}

\begin{table*} [ht!]
\small
\centering
\begin{tabular}{l|ccccccccc|c}
\toprule
\textbf{Model} & \textbf{IDC} & \textbf{PRAD} & \textbf{PAAD} & \textbf{SKCM} & \textbf{COAD} & \textbf{READ} & \textbf{ccRCC} & \textbf{LUNG} & \textbf{LYMPH IDC} & \textbf{Average} \\
\midrule
CONCH v15 & 0.504 & \third{0.373} & 0.391 & 0.460 & \second{0.264} & 0.157 & 0.188 & 0.501 & \third{0.257} & 0.344 \\
Phikon v2 & 0.536 & 0.303 & 0.410 & 0.495 & 0.221 & 0.146 & \first{0.250} & 0.470 & 0.224 & 0.339 \\
UNI v2 & \third{0.579} & 0.369 & 0.420 & 0.583 & 0.219 & 0.181 & 0.233 & 0.450 & 0.224 & 0.362 \\
H-optimus-0 & \second{0.586} & 0.327 & 0.434 & \first{0.609} & \third{0.257} & 0.192 & \second{0.239} & 0.495 & 0.227 & \third{0.374} \\
Prov-GigaPath & 0.537 & 0.346 & 0.398 & 0.524 & 0.227 & 0.155 & 0.208 & 0.460 & 0.211 & 0.341 \\
Virchow v2 & \first{0.589} & 0.351 & \first{0.457} & \third{0.599} & 0.242 & \third{0.207} & \third{0.237} & \second{0.559} & 0.242 & \second{0.387} \\
\midrule
Self-Attn & 0.571 & 0.340 & \second{0.454} & 0.596 & 0.214 & 0.203 & 0.225 & \first{0.562} & 0.239 & 0.378 \\
Top-3 MoE & 0.450 & 0.335 & 0.288 & 0.284 & 0.220 & 0.098 & 0.190 & 0.379 & 0.240 & 0.276 \\
AdaFusion-Coarse & 0.557 & \second{0.394} & 0.416 & 0.523 & 0.235 & \second{0.227} & 0.203 & 0.482 & \second{0.273} & 0.368 \\
AdaFusion-Fine & 0.565 & \first{0.407} & \third{0.442} & \second{0.600} & \first{0.279} & \first{0.240} & 0.236 & \third{0.542} & \first{0.277} & \first{0.399} \\
\bottomrule
\end{tabular}%
\caption{Detailed performance on the HEST-Benchmark regression task, reporting the Pearson Correlation Coefficient (PCC). PFMs are evaluated at their original feature dimensions, while all fusion methods use a 6$\times$64-dimensional representation. First, second, and third best results in each column are denoted as bold-underline, bold, and bold-italic.}
\label{tab:hest_detailed_results}
\end{table*}

\subsection{Datasets \& Experiments Setting} 
We evaluate our method on three public benchmarks spanning binary classification, multi-class classification, and regression tasks:
\begin{itemize}
    \item \textbf{ATEC23 (ATEC2023-Challenge)}~\cite{wang2025atec23}: An ovarian cancer dataset for predicting response to bevacizumab (effective vs. ineffective). WSIs are sourced from multiple centres. We use 5-fold cross-validation.

    \item \textbf{PANDA (Prostate cAncer graDe Assessment)}~\cite{bulten2022artificial}: A large-scale prostate cancer dataset with 10,616 WSIs annotated by ISUP grade (0–5). We perform 5-fold cross-validation on the development set.

    \item \textbf{HEST-Benchmark (Histology-based Gene Expression Signature Test)}~\cite{jaume2024hest}: A regression benchmark of nine tasks predicting expression levels of the top 50 spatially variable genes across cancer types. We follow the official training–testing splits.
\end{itemize}

\subsection{Implementation Details}
\noindent\textbf{Feature Extraction.} We use six PFMs to extract multi-level representations per tile: CONCH v1.5~\cite{lu2024visual}, Phikon v2~\cite{filiot2024phikon}, UNI v2~\cite{chen2024towards}, H-optimus-0~\cite{hoptimus0}, Prov-GigaPath~\cite{xu2024whole}, and Virchow v2~\cite{zimmermann2024virchow2}. Features are dimensionality-compressed and concatenated to form a unified tile embedding.

\noindent\textbf{Task Head.} For classification tasks (ATEC23 and PANDA), tile features are aggregated into slide-level representations via an Attention-based Multiple Instance Learning (ABMIL) model~\cite{ilse2018attention}, followed by a linear classifier. For regression (HEST-Benchmark), features are passed to a linear head to predict gene expressions.

\noindent\textbf{Training Setup.} All trainable modules (Prompt Tuner and task head) are optimised using Adam with a learning rate of 2e-4 and weight decay of 1e-5. Classification uses cross-entropy loss (50 epochs), and regression uses mean squared error (20 epochs). All models are trained on a single NVIDIA 4090 GPU.

\noindent\textbf{Evaluation Metrics.} We report Accuracy (ACC) and Area Under the ROC Curve (AUC) for classification tasks, and Pearson Correlation Coefficient (PCC) for regression. For cross-validation, mean scores are reported.

\noindent\textbf{Baselines.} We compare AdaFusion with every single PFM feature and two fusion baselines: (1) \textbf{Self-Attn} \cite{gao2025features}, which applies standard self-attention to concatenated features; and (2) \textbf{Top-3 MoE}, a Mixture-of-Experts model with a gating mechanism that selects the top three features for concatenation and remapping via a linear layer.

\begin{figure}[t!]
\centering
\includegraphics[width=0.45\textwidth]{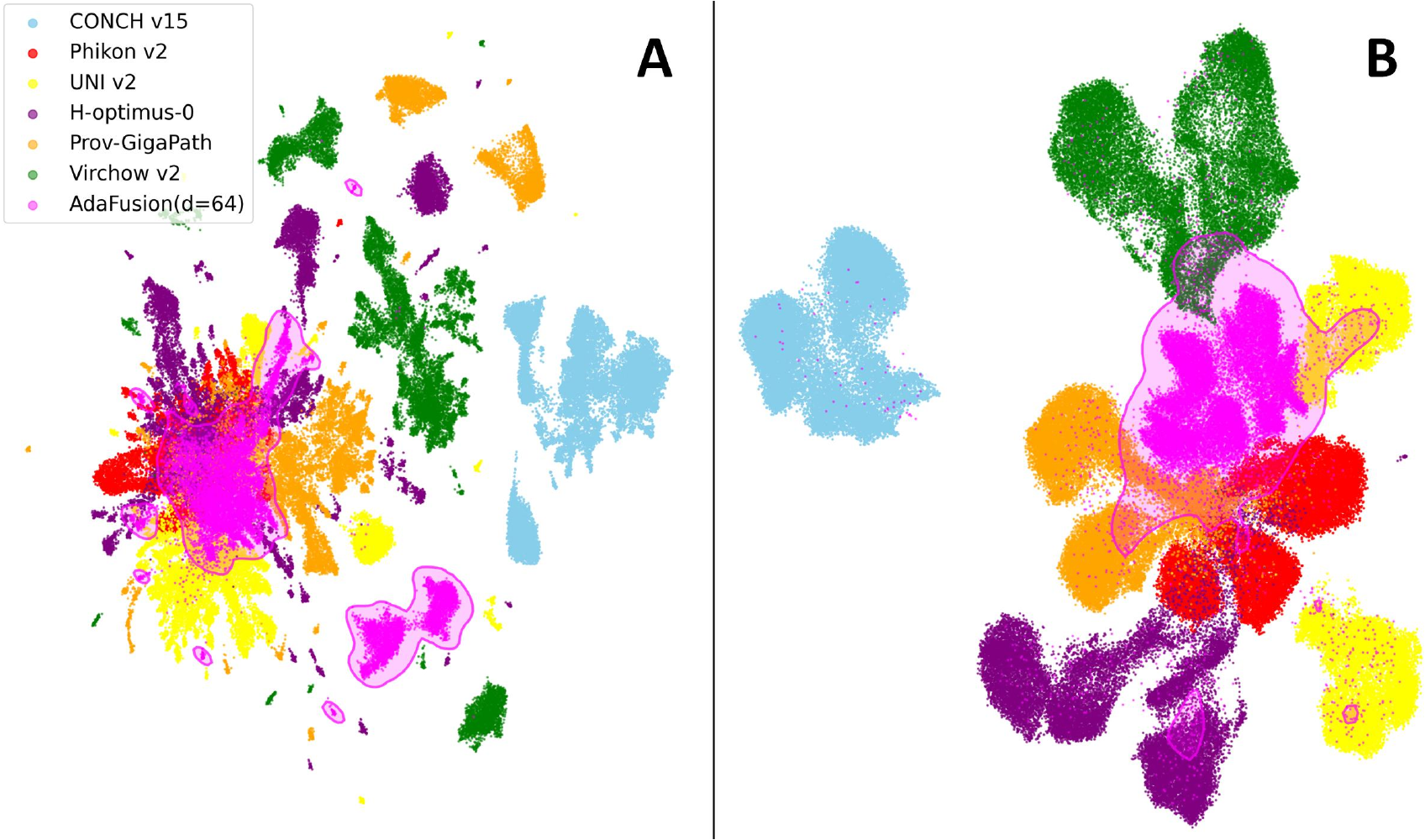}
\caption{
UMAP visualisation of tile features from the (A) ATEC23 and (B) PANDA datasets, extracted by PFMs and our AdaFusion-Fine. The features are from the top 10 most indicative tiles identified by an ABMIL model, which was trained based on ImageNet-pretrained ResNet50 tile-level features to ensure an unbiased selection basis.
}
\label{fig4}
\end{figure}

 \begin{figure*}[ht!]
\centering
\includegraphics[width=0.8\textwidth]{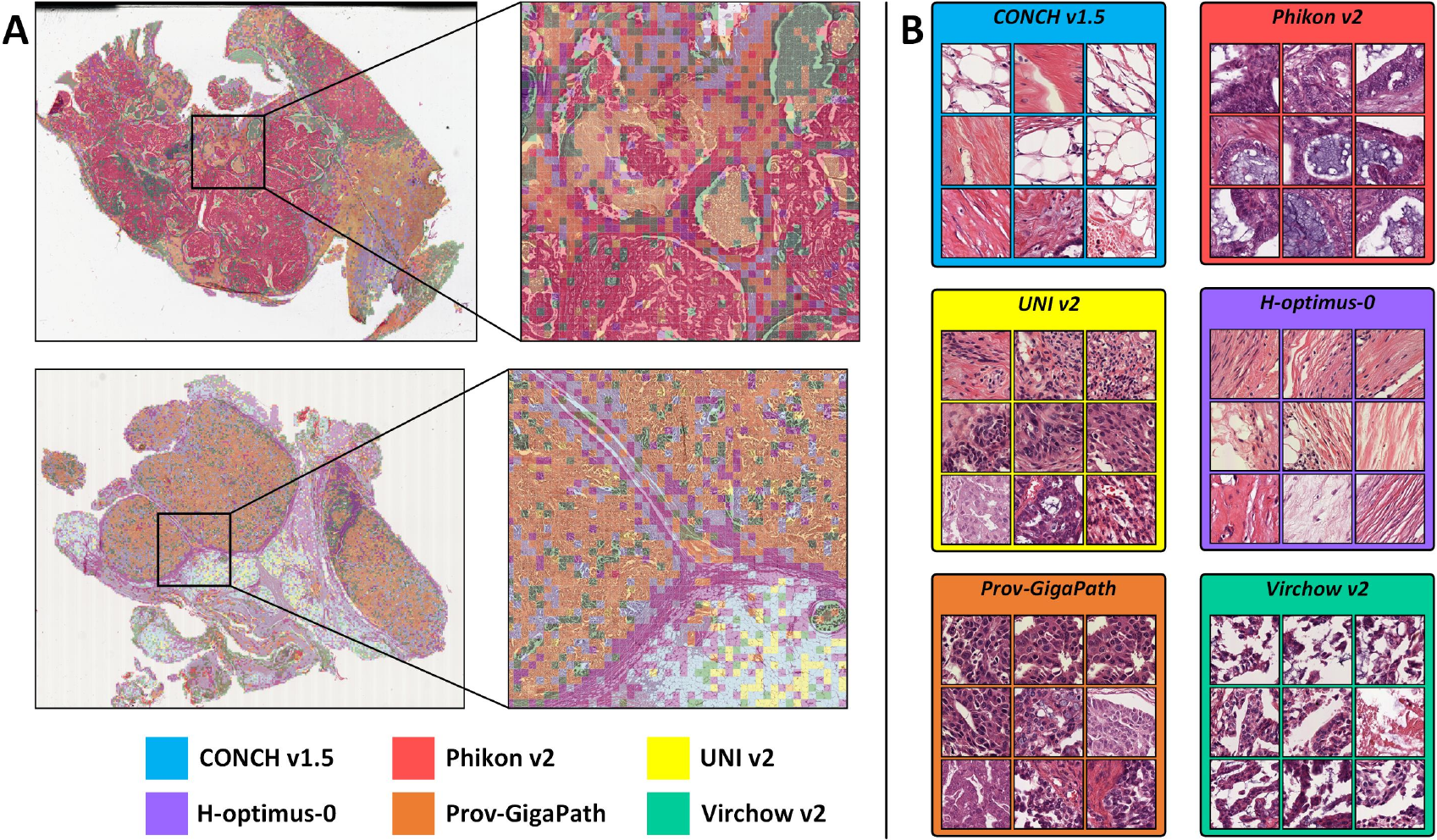}
\caption{
PFMs' contribution visualisation based on adaptive prompt-guided fusion (AdaFusion-Fine (d=64)). 
A. Each colour indicates the PFM with the highest contribution score across tissue regions. 
B. Representative tile examples most contributed by each model, showcasing their diverse morphological preferences.
}
\label{fig5}
\end{figure*}

\subsection{Experimental Results}
We present the comprehensive results of our evaluation in Table~\ref{tab:classification_results} and Table~\ref{tab:hest_detailed_results}. The experiments spanning both classification and regression tasks, consistently show that our proposed AdaFusion significantly outperforms individual state-of-the-art PFMs. This highlights the robustness and generalisability of fusing diverse feature representations.

\noindent\textbf{Results on Classification Tasks.}
Table~\ref{tab:classification_results} benchmarks our AdaFusion variants against individual PFM and established fusion baselines on the ATEC23 and PANDA datasets. The results consistently demonstrate the superiority of our proposed method across all evaluation settings.

On the ATEC23 dataset,  AdaFusion outperforms all PFMs and fusion baselines. In particular, the performance of our models remains robust even when feature dimensions are aggressively reduced. The compact 6$\times$64-dimensional versions achieve top-tier accuracy, matching higher-dimensional counterparts. The results highlight our method's exceptional ability to compress features and preserve essential information in a highly efficient, low-dimensional representation.

The PANDA multi-class grading task highlights the advantages of our approach more clearly.
Here, AdaFusion achieves higher performance under highly compressed 6$\times$64-dimensional features, significantly outperforming the comparison methods in terms of ACC and AUC. This validates our core hypothesis: AdaFusion effectively integrates complementary knowledge from different PFMs to create powerful and computationally efficient feature representations for complex classification tasks.

\noindent\textbf{Results on Regression Task.}
The effectiveness of our approach are further validated on the HEST-Benchmark dataset, with detailed results presented in Table~\ref{tab:hest_detailed_results}.

Overall, our method achieves the highest average Pearson Correlation Coefficient, decisively outperforming all individual models and the other fusion baselines. It demonstrates a clear advantage over the competitive Self-Attn method and a vast improvement over the Top-3 MoE, which struggles on this regression benchmark with low-dimensional features.

This aggregate superiority is driven by consistently dominant performance across the individual sub-tasks. AdaFusion-Fine secures the top rank in the majority of categories where it is not surpassed by a single, specialized PFM. Specifically, it achieves a top-three rank in seven of the nine sub-tasks. In stark contrast, no other fusion baseline secures a leading position on any sub-task, a result that unequivocally demonstrates the unique robustness and generalizability of AdaFusion's adaptive mechanism. 

In summary, the consistent, state-of-the-art performance across three distinct benchmarks strongly validates the effectiveness and robustness of our approach. By adaptively integrating representations from multiple PFMs, AdaFusion consistently surpasses not only individual models but also other powerful fusion techniques.

\subsection{Visualisation and Interpretability}
\noindent\textbf{Visualisation of Feature Fusion.}
To visually examine the feature fusion capability of our method, we present UMAP projections of tile-level features for both the ATEC23 and PANDA datasets in Figure~\ref{fig4}. To avoid representation bias from any single pre-trained model, we first trained an ABMIL model using features extracted from an ImageNet-pretrained ResNet50. From each slide, we then selected the 10 tiles with the highest attention scores for visualisation.

Figure~\ref{fig4} offers visual evidence of the effectiveness of our fusion mechanism. In Panel (A) for the ATEC23 dataset, features extracted by AdaFusion-Fine model occupy a central, integrative position. Instead of forming a distinct, separate cluster, the ensemble features are interspersed among the feature spaces of multiple individual models. This pattern is consistently observed in Panel (B) for the PANDA dataset. Despite the different underlying data distribution, the AdaFusion features again do not isolate themselves but rather co-locate with various constituent model clusters. This consistent behaviour across two distinct datasets strongly demonstrates that AdaFusion effectively learns a composite and robust feature space by integrating complementary information from its diverse source models, rather than simply creating another independent representation.

\noindent\textbf{Visualisation of Model Contributions.}
To further understand how our model dynamically leverages the knowledge from different PFMs, we visualise the contribution scores derived from our prompt-guided fusion mechanism in Figure~\ref{fig5}. Figure~\ref{fig5}A displays contribution masks overlaid on whole-slide images, where each colour denotes the PFM with the highest contribution score in the corresponding region. These colour-indicated areas align with distinct histomorphological phenotypes, suggesting that different models exhibit preferences for specific tissue structures (e.g., tumour, stroma, glandular areas). This specialisation is further validated in Figure~\ref{fig5}B, which displays representative tiles that correspond to the unique histological patterns most strongly favoured by each model. 
Collectively, these visualisations show that our method adaptively orchestrates the strengths of diverse PFMs, selectively applying them based on local pathological context. This not only contributes to improved performance, but also provides the natural and inherent interpretability for the idiosyncratic knowledge pattern of each PFM.

\begin{table}[ht!]
\small
\setlength{\tabcolsep}{1mm}
\centering
\begin{tabular}{l|cc|cc|c}
\toprule
\multicolumn{1}{c|}{\multirow{2}{*}{\textbf{Setting}}} & \multicolumn{2}{c|}{\textbf{ATEC23}} & \multicolumn{2}{c|}{\textbf{PANDA}} & \textbf{HEST} \\
\multicolumn{1}{c|}{} & ACC & AUC & ACC & AUC & PCC \\
\midrule
Ensemble  (d=64)   & 0.807 & 0.864 & 0.720 & 0.928 & 0.321 \\
Ensemble (d=64, w Mask)  & 0.841 & 0.873 & 0.743 & 0.938 & 0.336 \\
AdaFusion-Fine (d=64, full)   & \textbf{0.862} & \textbf{0.900} & \textbf{0.815} & \textbf{0.955} & \textbf{0.399} \\
\midrule
Ensemble (d=256)  & 0.855 & 0.895 & 0.752 & 0.942 & 0.379 \\
Ensemble (d=256, w Mask)  & 0.845 & 0.885 & 0.766 & 0.945 & 0.393 \\
AdaFusion-Fine (d=256, full)  & \textbf{0.863} & \textbf{0.909} & \textbf{0.817} & \textbf{0.956} & \textbf{0.423} \\
\midrule
Ensemble (d=512)  & 0.856 & 0.908 & 0.760 & 0.945 & 0.395 \\
Ensemble (d=512, w Mask)  & 0.855 & 0.909 & 0.777 & 0.950 & 0.400 \\
AdaFusion-Fine (d=512, full)  & \textbf{0.869} & \textbf{0.910} & \textbf{0.821} & \textbf{0.958} & \textbf{0.414} \\
\bottomrule
\end{tabular}%
\caption{Ablation study of our framework. We compare the full AdaFusion-Fine model against two baselines: direct feature concatenation (Ensemble) and concatenation with a random mask (Ensemble w Mask). Experiments are run with PFM's features pooled to different pre-concatenation dimensions ($d$). Bold numbers indicate the best performance.}
\label{tab:ablation_method}
\end{table}

\subsection{Ablation Studies}
To validate our design choices and understand the source of performance gains, we conduct a series of ablation studies focusing on the contribution of our core components.

\noindent\textbf{Impact of the Prompt Tuning Module.}
We perform an ablation study to isolate the effectiveness of our adaptive fusion strategy. We compare our full model, AdaFusion-Fine, against two progressively stronger baselines:
\begin{itemize}
    \item \textbf{Ensemble:} Features from all PFMs are directly concatenated and fed to the downstream task head. This represents, a simple fusion approach.
    \item \textbf{Ensemble (w Mask):} A random masking strategy is applied to the concatenated features before the task head, serving as a simple regularisation technique.
\end{itemize}
These comparisons are evaluated across various feature dimensions ($d \in \{64, 256, 512\}$), where $d$ is the dimension of each PFM's feature vector prior to concatenation.

As shown in Table~\ref{tab:ablation_method}, our full AdaFusion-Fine model consistently and substantially outperforms both baselines across all datasets and dimensionalities. The results clearly demonstrate that a sophisticated, adaptive fusion mechanism is superior to simple concatenation or regularisation. These significant gains validate that our prompt tuning module is not merely aggregating features but is effectively learning to recalibrate and synergise heterogeneous information for specific downstream tasks.

Furthermore, while applying a random mask sometimes offers a slight improvement over the basic ensemble, it consistently falls short of the performance achieved by our adaptive approach. This underscores that the learned, sample-specific reweighting from our prompt tuner is far more effective than a simple, stochastic regularisation. The robust, state-of-the-art performance of AdaFusion across all tested dimensions also highlights its low sensitivity to this hyperparameter, further validating its practical utility.

\section{Conclusion}
\label{sec5}
In this work, we introduced \textbf{AdaFusion}, a prompt-guided framework for effectively integrating knowledge from multiple, diverse PFMs. By leveraging a lightweight, trainable prompt tuner over low-dimensional, frozen features, AdaFusion consistently outperforms not only individual PFMs but also established fusion baselines across challenging classification and regression benchmarks. The core strength of our approach is its parameter efficiency and inherent interpretability, which enables quantitative analysis of each PFM's contribution based on histopathological context. Future directions include automating PFM ensemble selection and deepening the link between model contributions and specific histological patterns. Ultimately, AdaFusion represents a significant step towards more robust, efficient, and transparent multi-PFM systems in computational pathology.

\bibliography{ref}


\clearpage

\appendix
\section{APPENDIX}

\begin{table*}[ht!]
\centering
\small
\begin{tabular}{p{2.5cm} p{3.0cm} p{4.5cm} p{4.5cm}}
\toprule
\textbf{Model} & \textbf{Architecture} & \textbf{Pre-training Data} & \textbf{Pre-training Method}\\
\midrule
\textbf{CONCH v15} & 
 Vision-Language (ViT-B+PubMedBERT) & 
 1.17M multi-modal image-text pairs. &
 Contrastive V-L alignment and captioning. \\
\midrule
\textbf{Phikon v2} & 
 ViT-L & 
 Public TCGA dataset, focused on primary tumours. &
 iBOT (Masked Image Modelling). \\
\midrule
\textbf{UNI v2} & 
 ViT-L &
 Large-scale (100K WSIs) but from a single institutional system (Mass General Brigham). & 
 DINOv2 (Contrastive / Self-Distillation). \\
\midrule
\textbf{H-optimus-0} & 
 ViT-G &
 Massive (500K WSIs) but fully proprietary and opaque data source. & 
 DINOv2 (Contrastive / Self-Distillation). \\
\midrule
\textbf{Prov-GigaPath} &
 ViT-G tile encoder + LongNet slide encoder. &
 Large-scale (171K WSIs) from a single US health network (Providence). &
 Two-stage: Tile-level DINOv2, then slide-level MAE. \\
\midrule
\textbf{Virchow v2} &
 ViT-H &
 Very large (3.1M WSIs) from a major cancer centre (MSKCC) and partners. Mixed magnifications. & 
 Advanced DINOv2 (improved regularisation and augmentation). \\
\bottomrule
\end{tabular}
\caption{Comparison of six PFMs in terms of architecture, data, and training approach.}
\label{tab:pfm_deep_analysis}
\end{table*}

\begin{table}[ht!]
\centering
\small
\setlength{\tabcolsep}{1mm}
\begin{tabular}{lcccc}
\toprule
\textbf{Model / Method} & \multicolumn{4}{c}{\textbf{FPS}} \\
\cmidrule(lr){2-5}
 & \textbf{(Native Dim)} & \textbf{d=64} & \textbf{d=256} & \textbf{d=512} \\
\midrule
\multicolumn{5}{l}{\textit{\textbf{A) PFMs}}} \\
\quad CONCH v15         & 36.42 & - & - & - \\
\quad Phikon v2         & 29.74 & - & - & - \\
\quad UNI v2            & 24.42 & - & - & - \\
\quad H-optimus-0       & 24.42 & - & - & - \\
\quad Prov-GigaPath     & 24.42 & - & - & - \\
\quad Virchow v2        & 14.59 & - & - & - \\
\midrule
\multicolumn{5}{l}{\textit{\textbf{B) Fusion Methods}}} \\
\quad Self-Attn         & -      & 16.47 & 4.32  & 2.20  \\
\quad Top3-MoE          & -      & 53.90 & 21.66 & 12.94 \\
\quad AdaFusion-Coarse & - & 52.41 & 21.26 & 11.22 \\
\quad AdaFusion-Fine   & - & 52.37 & 17.91 & 7.36  \\
\bottomrule
\end{tabular}
\caption{Inference speed of all benchmarked models.}
\label{tab:fps_results_unified}
\end{table}

\begin{figure*}[ht!]
    \centering
    \includegraphics[width=\textwidth]{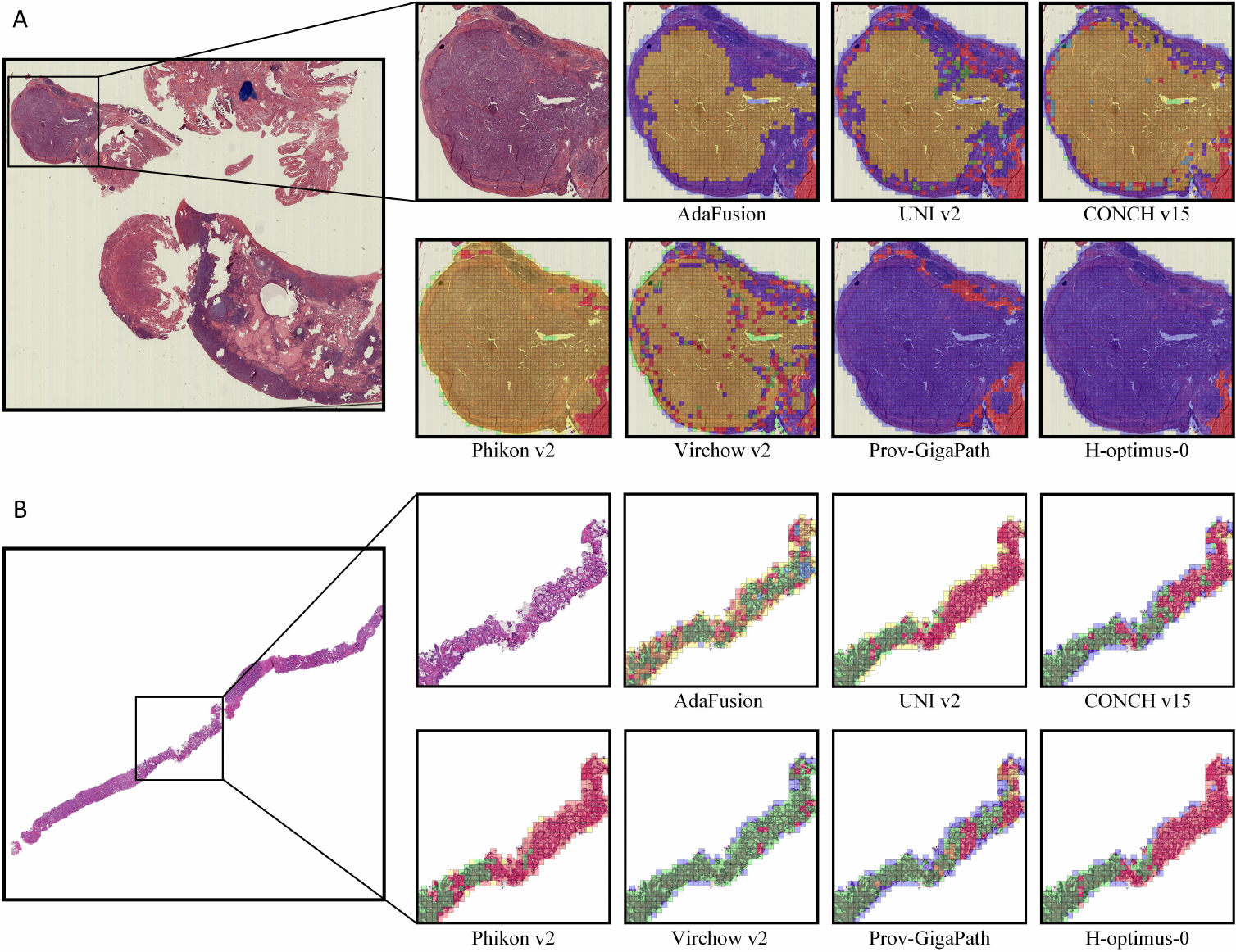}
    \caption{Clustering analysis visualisation of features from AdaFusion and each of PFMs on (A) ATEC23 and (B) PANDA datasets. K-Means clustering ($k=5$) reveals AdaFusion’s capacity to delineate different histomorphological phenotypes.}
    \label{fig6}
\end{figure*}

\begin{figure*}[ht!]
\centering
\includegraphics[width=\textwidth]{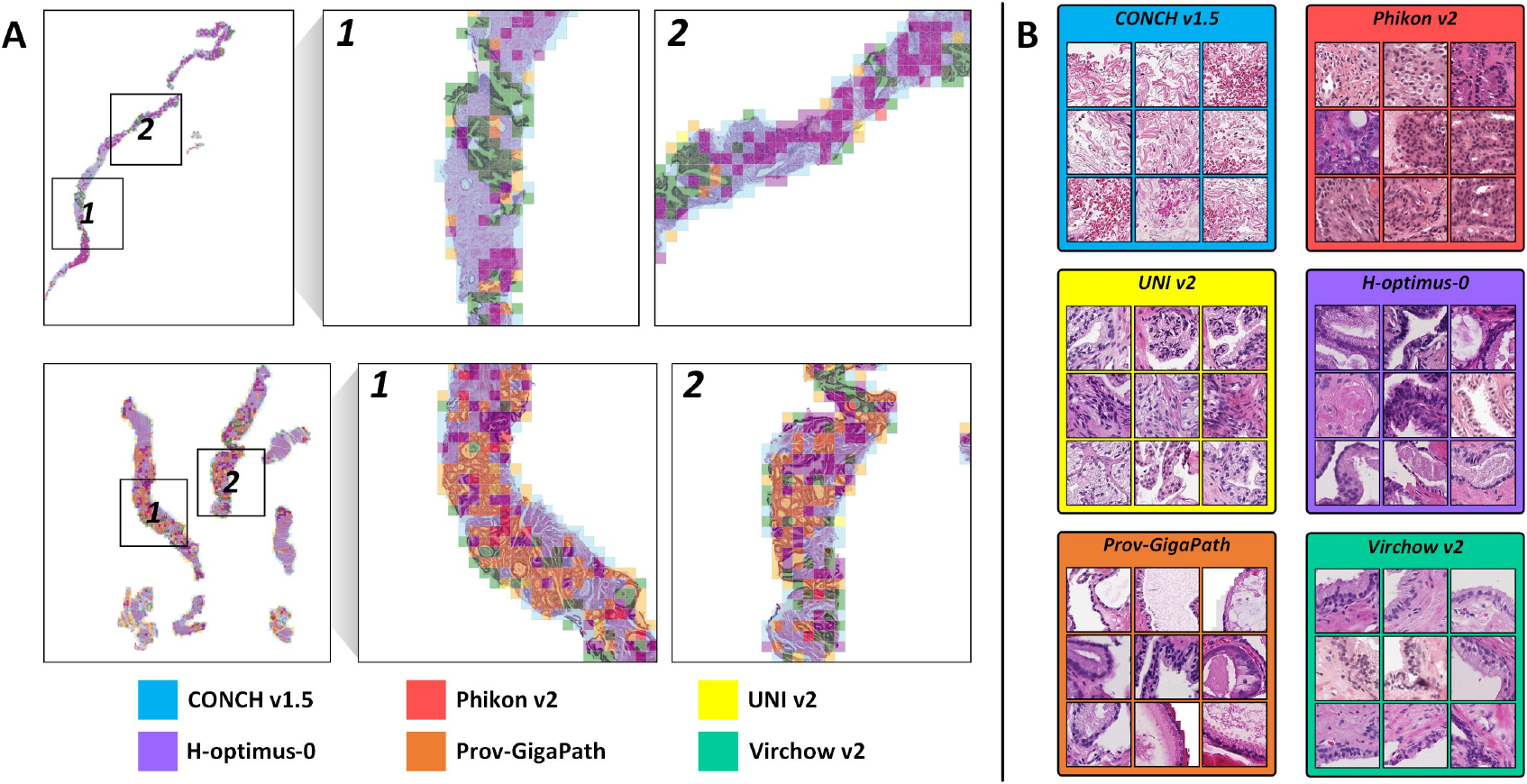} 
\caption{PFM contribution visualisation on the PANDA dataset using AdaFusion-Fine ($d=64$).
A. Contribution masks, where each colour indicates the PFM with the highest contribution across tissue regions.
B. Representative tiles most contributed by each model, illustrating their distinct morphological preferences.}
\label{fig7}
\end{figure*}

\subsection{A. Implementation Details of Baselines}
\label{appendix:baselines}

\noindent We provide implementation details of the two baseline methods used for comparison. Both baselines take pooled and concatenated features from all PFMs as input and are trained end-to-end with the downstream task head.

\noindent\textbf{Self-Attn Baseline.}
This baseline consists of a standard Transformer encoder block to model dense, all-to-all interactions among concatenated feature dimensions.

\begin{itemize}
    \item \textbf{Input Dimension:} Concatenated feature vector $\mathbf{E}_{\text{compound}} \in \mathbb{R}^{N \times d}$, where $N \times d$ depends on the setup (e.g., $d = 6 \times 64 = 384$ for 64-dimensional experiments).
    
    \item \textbf{Multi-Head Self-Attention (MHSA):}
        \begin{itemize}
            \item Number of Attention Heads: 8.
            \item Head Dimension: Each head has dimension $N \times d/8$.
        \end{itemize}
    
    \item \textbf{Feed-Forward Network (FFN):} Follows the MHSA block and consists of two linear layers with a GELU activation in between.
        \begin{itemize}
            \item Hidden Dimension: $4 \times N \times d$.
        \end{itemize}
\end{itemize}

\noindent \textbf{Top-3 MoE Baseline.}
This baseline implements a sparse Mixture-of-Experts (MoE) design. A gating network scores the relevance of each PFM “expert”, selects the top three, and combines their features.

\begin{itemize}
    \item \textbf{Input Dimension:} Concatenated feature vector $\mathbf{E}_{\text{compound}} \in \mathbb{R}^{N \times d}$, where $d$ is the feature dimension of a single PFM and $N$ is the number of PFMs.
    
    \item \textbf{Gating Network:} A two-layer MLP that maps the input to expert relevance scores.
        \begin{itemize}
            \item Architecture: Two linear layers with GELU activation; mapping $\mathbb{R}^{N \times d} \to \mathbb{R}^{d} \to \mathbb{R}^{K}$.
            \item Activation: Softmax applied to generate probability weights over experts.
        \end{itemize}
    
    \item \textbf{Expert Selection:} A non-differentiable Top-K operation selects indices of the three highest-scoring experts, enabling hard routing.
    
    \item \textbf{Fusion by Concatenation:} Selected feature vectors (each $\in \mathbb{R}^{d}$) are concatenated to form a fused representation.
        \begin{itemize}
            \item Output Dimension: $\mathbb{R}^{3 \times d}$.
        \end{itemize}
    
    \item \textbf{Final Projection:} A linear layer remaps the fused vector to a fixed-size representation before prediction. Mapping: $\mathbb{R}^{3 \times d} \to \mathbb{R}^{3 \times d}$.
\end{itemize}

\subsection{B. Supplementary Analysis of PFM Heterogeneity}
\label{appendix:pfm_biases}

\noindent A central motivation of AdaFusion is that leveraging diverse PFMs may help reduce individual model biases and lead to more robust, generalisable representations. This section provides additional context on the heterogeneity across PFMs, summarising their distinct characteristics in terms of training data, architecture, and learning objectives.

As discussed in the introduction, biases in PFMs may arise from two main sources:

\noindent \textbf{Data-Induced Biases:} These stem from the composition of pretraining datasets. Examples include the use of non-public or institution-specific data, limited geographic or demographic coverage, and uneven distributions of cancer types or tissue representations.

\noindent \textbf{Architectural and Methodological Biases:} These relate to the design and training approach of each model. Key aspects include:
\begin{itemize}
    \item \textbf{Architectural Focus:} Vision Transformers (ViT) are well-suited for modelling local, patch-level features, while dilated or long-range attention architectures (e.g., LongNet) are designed to capture broader tissue context.
    
    \item \textbf{Learning Objective:} Contrastive methods (e.g., DINOv2) promote separability between instances by enforcing discriminative representations. Masked autoencoding (e.g., MAE) encourages learning holistic, context-aware features through reconstruction. Vision–Language (V–L) models align visual features with text-based semantics, offering language-informed representations.
\end{itemize}

Such differences may result in PFMs with complementary strengths and weaknesses. For instance, models trained on TCGA data may generalise well to prevalent cancer types but underperform on rare subtypes, while V–L models might identify high-level linguistic categories but overlook subtle or tacit morphological cues. 

Table~\ref{tab:pfm_deep_analysis} summarises the heterogeneity of the six PFMs used in our study, covering their data sources, architectural choices, and training algorithms. This landscape of complementary biases supports the motivation for AdaFusion, which is designed to act as a dynamic “coordinator”, adaptively adjusting the influence of each PFM according to the tissue context of each sample.

\subsection{C. Computational Efficiency}
\label{appendix:efficiency}

\noindent Beyond predictive performance, the computational efficiency of AdaFusion is a key consideration for practical deployment. To quantify this, we benchmarked its inference speed against individual PFMs and other fusion baselines. Inference speed is reported in Frames Per Second (FPS), where one “frame” corresponds to one WSI.

\noindent \textbf{Experimental Setup. } We simulated 20 synthetic feature sets, each representing a WSI with tile counts ranging from 2,000 to 40,000 (in steps of 2,000). Binary classification inference was then performed using ABMIL-aggregated features. For each configuration, the average FPS was measured over multiple runs on a single NVIDIA 4090 GPU. Results are summarised in Table~\ref{tab:fps_results_unified}.

\noindent \textbf{Analysis. } The results highlight several key findings:
\begin{itemize}
    \item \textbf{Efficiency of AdaFusion:} AdaFusion, particularly the fine variant with reduced feature dimensions, achieves significantly higher FPS than most individual PFMs and remains comparable to lightweight fusion baselines such as Top-3 MoE.

    \item \textbf{Cost of Self-Attention:} The Self-Attn baseline exhibits sharp degradation in FPS as feature dimensionality increases, indicating poor scalability for high-dimensional fusion scenarios.

    \item \textbf{Speed–Performance Trade-off:} AdaFusion-Fine ($d=64$) achieves strong predictive performance while maintaining over 2$\times$ the inference speed of most full-size PFMs. Among fusion methods, it outperforms Self-Attn in accuracy while delivering at least 3$\times$ higher throughput than this competitive baseline. This demonstrates its favourable balance of predictive power and computational efficiency for large-scale WSI deployment.
\end{itemize}

\subsection{D. Representative Capacity of Fused Features}
\label{appendix:cluster_qualitative}

\noindent We investigate whether the fused features offer more refined representations of morphological phenotypes compared to those from individual PFMs, using clustering analysis as illustrated in Figure~\ref{fig6}. We apply K-Means clustering ($k=5$) on tile-level features extracted by AdaFusion-Fine ($d=64$), which is the variant used throughout our visualisation studies, as well as by individual PFMs, across WSIs from the ATEC23 (Panel A) and PANDA (Panel B) datasets. Each colour in the resulting heatmaps corresponds to a distinct cluster ID, enabling assessment of the feature space’s separability and correspondence with tissue morphology.

In Panel A, we examine a heterogeneous tissue region from ATEC23 containing tumour and stromal components. AdaFusion produces clearly defined, spatially coherent clusters, suggesting a good capacity to delineate distinct morphological regions. By contrast, feature spaces from individual models such as UNI v2 and Virchow v2 exhibit mixed or fragmented cluster boundaries, implying their features' weaker alignment with underlying histological structure.

Panel B presents a WSI from PANDA, where accurate grading requires distinguishing subtle Gleason pattern variations. In this example, AdaFusion reveals a finer ability to characterise the tissue microenvironment. Rather than collapsing all malignant areas into a single broad category, the fused features capture intra-class variability, distinguishing and separating sub-regions that lie close together spatially yet differ in morphological phenotype.

These visualisations provide an additional line of evidence for the advantages of AdaFusion. The ATEC23 result underscores its ability to separate heterogeneous phenotypes, while the PANDA example highlights its strength in capturing subtle distinctions within similar histological phenotypes.

\subsection{E. Extended Visualisations of Model-Specific Contributions}
\label{appendix:more_visualizations}

\noindent In the main text, we visualised PFM-specific contribution maps to illustrate the interpretability of AdaFusion’s dynamic fusion mechanism. This section presents additional results on the ATEC23 and PANDA datasets to further demonstrate the alignment between model preference and histomorphological structures.

\noindent \textbf{Visual Analysis on the ATEC23 Dataset. } Figure~\ref{fig7} shows extended contribution maps for diverse regions in ATEC23. The colours indicate which PFM contributes most to each tile. These maps reveal clear spatial organisation aligned with tissue morphology, rather than random or uniform distributions. This supports our hypothesis that PFMs, shaped by their distinct training contexts, exhibit preferences towards specific histological phenotypes. For example, Phikon v2 (red) frequently dominates in densely cellular tumour areas, while Virchow v2 (green) and Prov-GigaPath (orange) are more active in stromal and connective tissue regions.

\begin{figure*}[ht!]
\centering
\includegraphics[width=\textwidth]{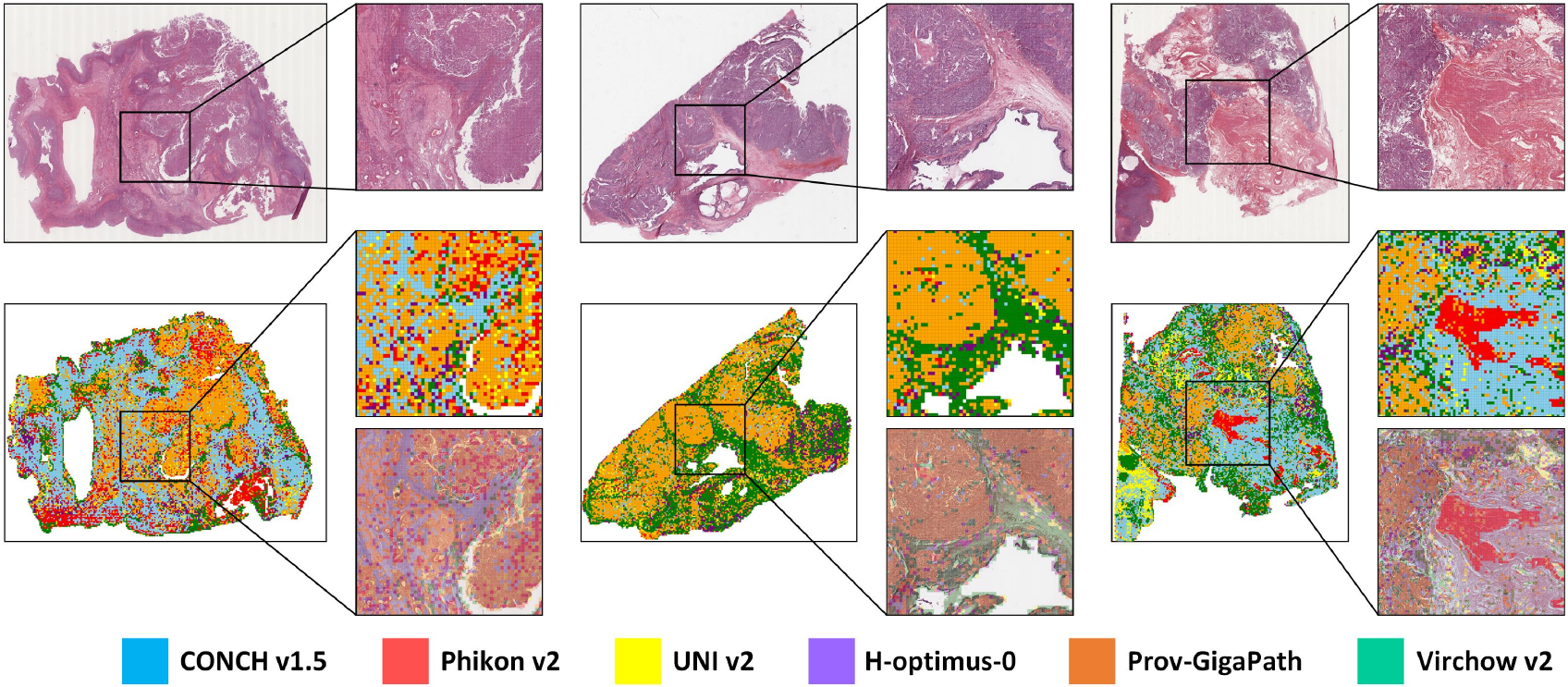} 
\caption{More visualisation examples of PFM contributions on the ATEC23 dataset. For three different WSI samples, we show the original H\&E image, the raw contribution mask, and a close-up overlay.
}
\label{fig:atec_vis}
\end{figure*}

\noindent \textbf{Visual Analysis on the PANDA Dataset. } Figure~\ref{fig:atec_vis} presents supplementary visualisations for the PANDA dataset. The model-specific contribution patterns and representative tiles further highlight the morphological specialisation across PFMs. For instance, Phikon v2 and UNI v2 tend to focus on hypercellular tumour regions, whereas CONCH v1.5 consistently aligns with fibrous stromal areas. H-optimus-0 and Virchow v2 display strong preferences for glandular structures, capturing different spatial cues relevant to the downstream task - Gleason grading.

\begin{figure*}[ht!]
\centering
\includegraphics[width=\textwidth]{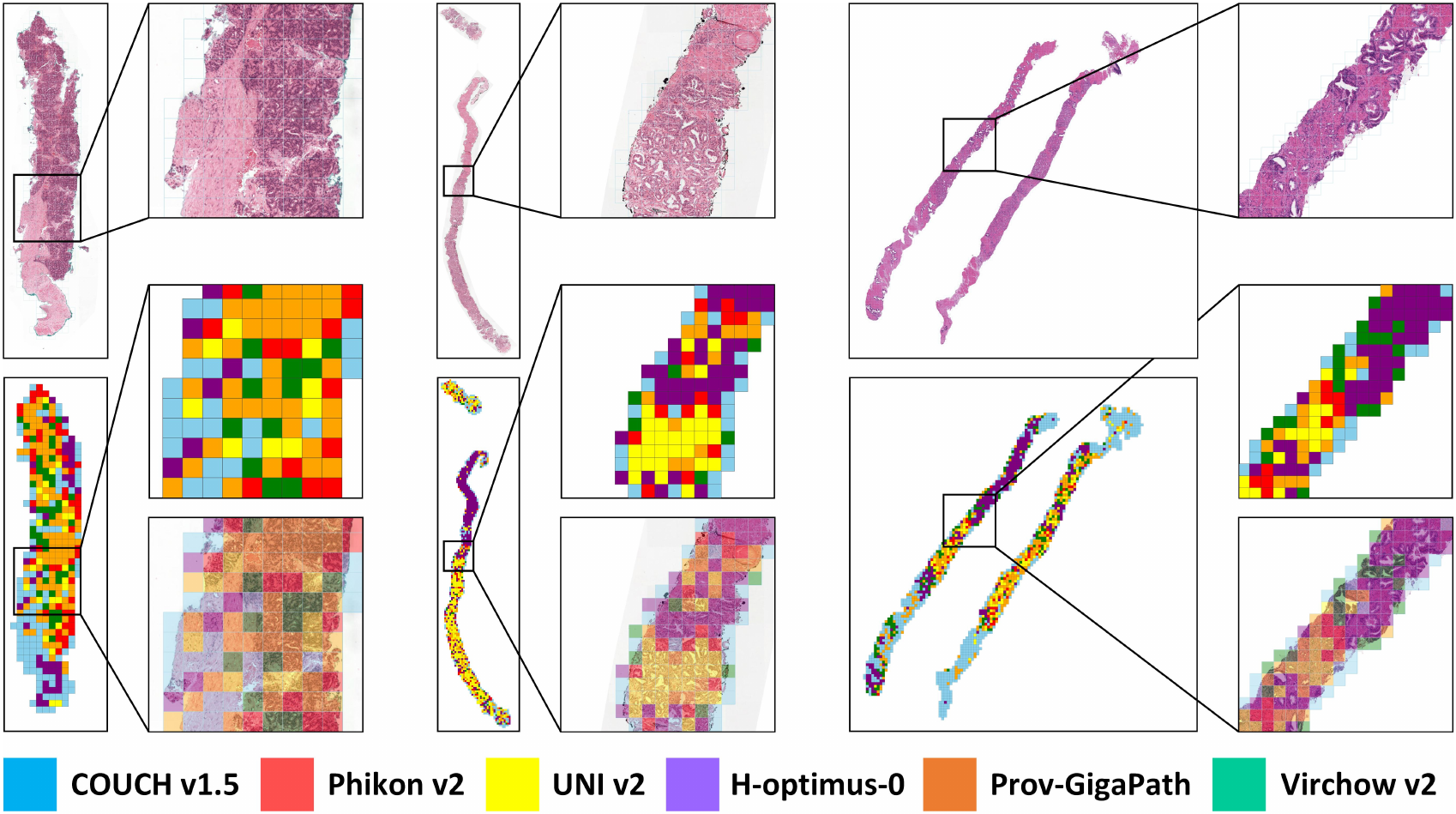}
\caption{More PFM contribution visualisations on the PANDA dataset. Same as above, for three WSI samples, we show the H\&E image, raw contribution mask, and close-up overlay.
}
\label{fig9}
\end{figure*}

\noindent Figure~\ref{fig9} offers more examples from PANDA, providing raw contribution maps and transparency overlays on the original H\&E images. These recurrent, interpretable preference patterns underscore AdaFusion’s ability to modulate expert selection based on histological context, leveraging each PFM’s internal biases to assign specialised roles across morphological phenotypes.

\begin{table*}[htbp]
\centering
\small
\setlength{\tabcolsep}{1.8mm}
\begin{tabular}{l|ccccccccc|c}
\toprule
\textbf{Model} & \textbf{IDC} & \textbf{PRAD} & \textbf{PAAD} & \textbf{SKCM} & \textbf{COAD} & \textbf{READ} & \textbf{ccRCC} & \textbf{LUNG} & \textbf{LYMPH IDC} & \textbf{Average} \\
\midrule
CONCH v15 & 0.504 & 0.373 & 0.391 & 0.460 & 0.264 & 0.157 & 0.188 & 0.501 & 0.257 & 0.344 \\
Phikon v2 & 0.536 & 0.303 & 0.410 & 0.495 & 0.221 & 0.146 & \first{0.250} & 0.470 & 0.224 & 0.339 \\
UNI v2 & 0.579 & 0.369 & 0.420 & 0.583 & 0.219 & 0.181 & 0.233 & 0.450 & 0.224 & 0.362 \\
H-optimus-0 & 0.586 & 0.327 & 0.434 & 0.609 & 0.257 & 0.192 & \third{0.239} & 0.495 & 0.227 & 0.374 \\
Prov-GigaPath & 0.537 & 0.346 & 0.398 & 0.524 & 0.227 & 0.155 & 0.208 & 0.460 & 0.211 & 0.341 \\
Virchow v2 & 0.589 & 0.351 & 0.457 & 0.599 & 0.242 & 0.207 & 0.237 & 0.559 & 0.242 & 0.387 \\
\midrule
Self-Attn(d=64) & 0.571 & 0.340 & 0.454 & 0.596 & 0.214 & 0.203 & 0.225 & 0.562 & 0.239 & 0.378 \\
Top3-MoE(d=64) & 0.450 & 0.335 & 0.288 & 0.284 & 0.220 & 0.098 & 0.190 & 0.379 & 0.240 & 0.276 \\
AdaFusion-Coarse(d=64) & 0.557 & 0.394 & 0.416 & 0.523 & 0.235 & 0.227 & 0.203 & 0.482 & \second{0.273} & 0.368 \\
AdaFusion-Fine(d=64) & 0.565 & 0.407 & 0.442 & 0.600 & 0.279 & \third{0.240} & 0.236 & 0.542 & \first{0.277} & 0.399 \\
\midrule
Self-Attn(d=256) & 0.558 & 0.360 & 0.455 & 0.648 & 0.213 & 0.192 & 0.162 & 0.554 & 0.185 & 0.369 \\
Top3-MoE(d=256) & 0.539 & 0.370 & 0.425 & 0.515 & 0.272 & \second{0.251} & 0.235 & 0.535 & 0.242 & 0.376 \\
AdaFusion-Coarse(d=256) & \second{0.595} & \second{0.419} & \third{0.480} & 0.612 & \second{0.287} & 0.230 & 0.235 & 0.561 & 0.268 & 0.410 \\
AdaFusion-Fine(d=256) & 0.594 & \first{0.421} & \first{0.497} & \third{0.654} & \first{0.293} & 0.237 & \second{0.249} & \first{0.594} & 0.270 & \first{0.423} \\
\midrule
Self-Attn(d=512) & 0.533 & 0.271 & 0.438 & 0.646 & 0.199 & 0.187 & 0.140 & 0.554 & 0.160 & 0.347 \\
Top3-MoE(d=512) & 0.567 & 0.344 & 0.454 & 0.546 & 0.216 & 0.227 & 0.222 & 0.572 & 0.246 & 0.377 \\
AdaFusion-Coarse(d=512) & \first{0.596} & \third{0.408} & \first{0.497} & \second{0.667} & \third{0.284} & \first{0.253} & 0.236 & \third{0.578} & \third{0.271} & \second{0.421} \\
AdaFusion-Fine(d=512) & \third{0.594} & \third{0.408} & \second{0.492} & \first{0.668} & 0.282 & 0.208 & 0.233 & \second{0.590} & 0.250 & \third{0.414} \\
\bottomrule
\end{tabular}
\caption{Complete performance on the HEST-Benchmark regression task, reporting the Pearson Correlation Coefficient (PCC). The table compares all variants of AdaFusion at different dimensions against the six individual PFMs. First, second, and third best results in each column are denoted as bold-underline, bold, and bold-italic.}
\label{tab:hest_full_results}
\end{table*}

\begin{figure*}[!t]
    \centering
    \includegraphics[width=\textwidth]{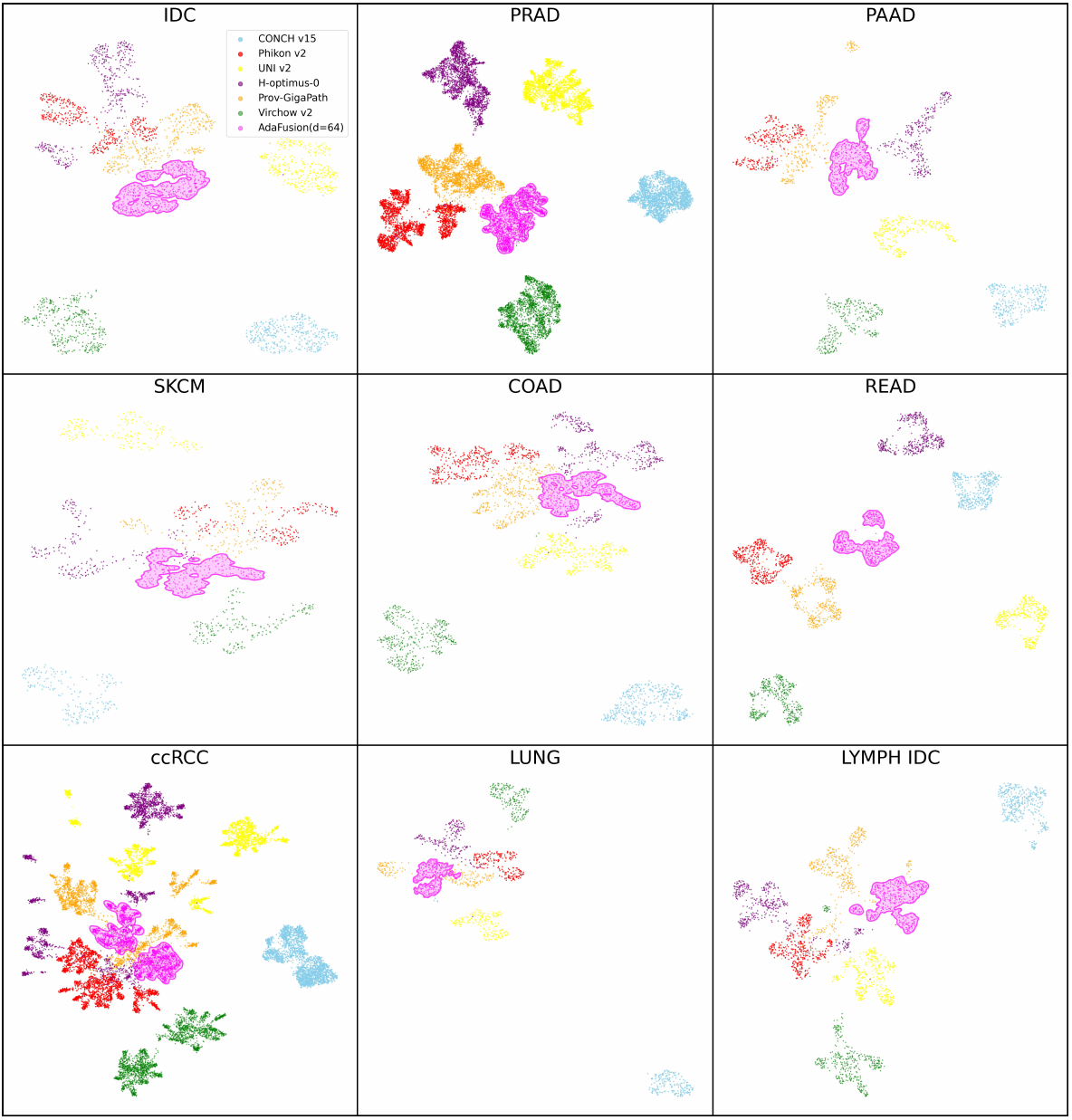}
    \caption{UMAP visualisation of tile-level feature spaces for the nine sub-tasks within the HEST-Benchmark.}
    \label{fig10}
\end{figure*}

\begin{figure*}[ht!]
\centering
\includegraphics[width=\textwidth]{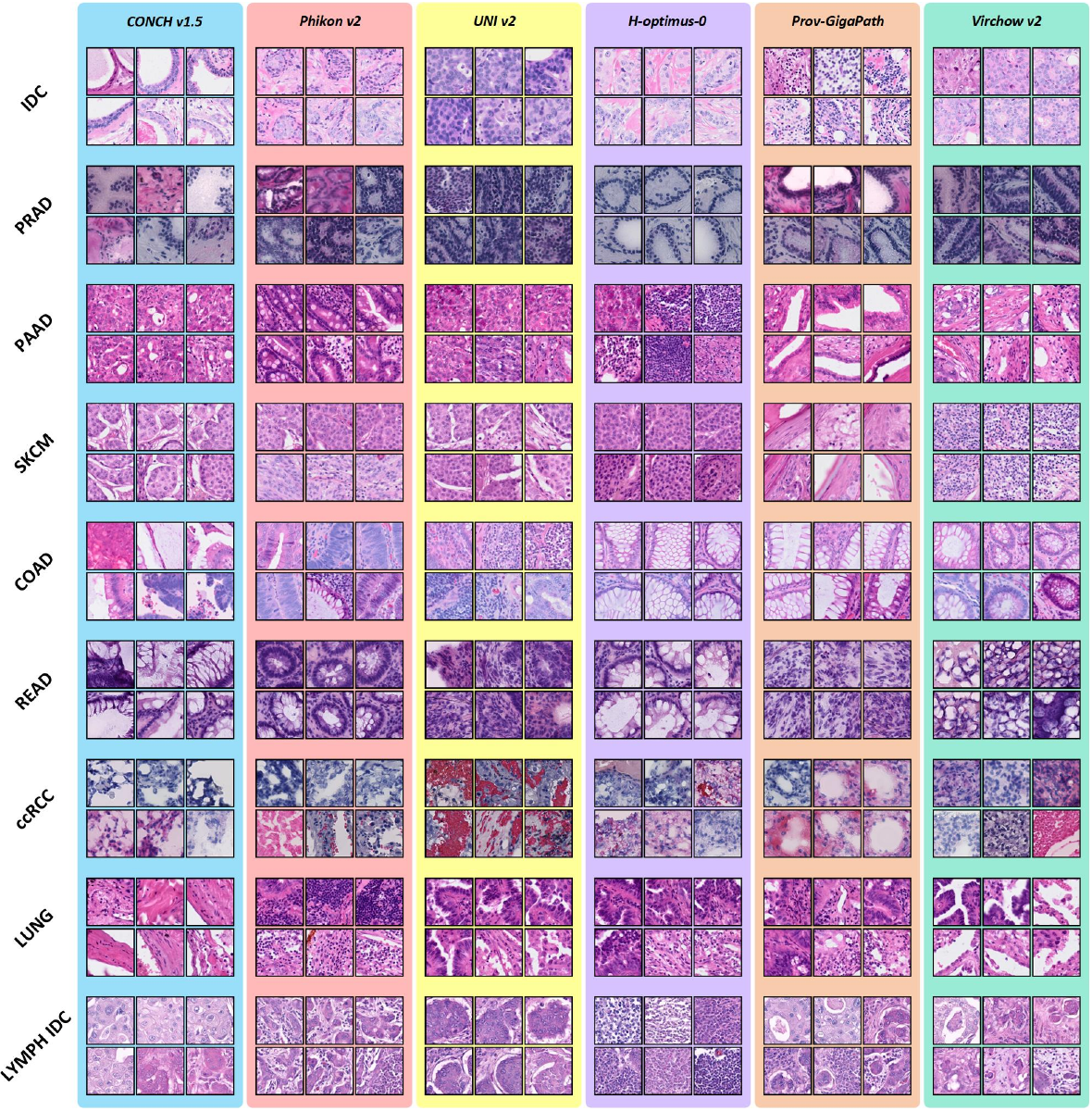} 
\caption{Visualisation of PFM morphological specialisations across the HEST-Benchmark. Each row corresponds to a different cancer type (sub-task), and each column represents one of the six PFMs.
}
\label{fig11}
\end{figure*}

\subsection{F. Complete Results on HEST Benchmark}
\label{appendix:hest_results}

\noindent Table~\ref{tab:hest_full_results} reports the complete Pearson Correlation Coefficient (PCC) results on the HEST regression benchmark, including all nine sub-tasks. It compares baseline methods, AdaFusion-Coarse and AdaFusion-Fine variants across feature dimensions of 64, 256, and 512, alongside the original performance of six individual foundation models at their native dimensions.

\noindent The results reinforce the conclusions in the main text, and several key patterns emerge:
\begin{itemize}
    \item \textbf{Overall Superiority:} Both AdaFusion variants consistently rank among the top three performers across nearly all sub-tasks, as well as in the overall average score. This highlights the advantage of our adaptive fusion framework over any individual PFM.

    \item \textbf{Comparison of AdaFusion Variants:} AdaFusion-Fine generally delivers the highest performance, exemplified by its leading average score at the 256-dimensional setting. AdaFusion-Coarse, meanwhile, serves as a highly competitive and more computationally efficient alternative, achieving the best results on specific sub-tasks such as IDC and READ at 512 dimensions.

    \item \textbf{Effect of Dimensionality:} The results suggest a clear performance increase from 64 to 256 dimensions, with diminishing returns beyond that point. Notably, even the compact AdaFusion-Fine (d=64) outperforms the strongest individual model, Virchow v2, indicating that our fusion mechanism effectively mitigates information redundancy and retains essential morphological signals under low-dimensional constraints.
\end{itemize}

\subsection{G. Feature Space Visualisation on HEST Benchmark}

We supplement our UMAP visualisation analysis for HEST-Benchmark, which comprises nine distinct regression sub-tasks across multiple cancer types. Following the same procedure used in the main paper, we randomly selected 100 tile features from each sub-task for visualisation.

The results, illustrated in Figure~\ref{fig10}, align closely with those observed in the ATEC23 and PANDA classification tasks. Across nearly all nine sub-tasks, features generated by AdaFusion occupy a central and integrative position within the collective feature space. Importantly, these fused features do not form a distinct, marginalised cluster; instead, they tend to occupy intermediate positions between the feature distributions of individual PFMs, bridging the representational gaps across models.

\subsection{H. Visualisation of PFM Specialisations on HEST Benchmark}
\label{appendix:hest_specialization}

\noindent We visualise tile-level samples across the nine sub-tasks of the HEST benchmark, as shown in Figure~\ref{fig11}. Each tile is attributed to a particular PFM based on the highest contribution score produced by the AdaFusion gating mechanism. This enables us to compile a visual library highlighting the preferred histological phenotypes for each foundation model.

The visualisations demonstrate that AdaFusion effectively highlights each PFM's distinct preferences towards specific histological phenotypes. For instance, Phikon v2 and UNI v2 consistently focus on regions characterised by high cellularity, dense nuclei, and poorly differentiated tumour cells across diverse cancer types such as IDC, SKCM, and LUNG, which indicating robust and generalisable specialisation in detecting aggressive or proliferative tumour patterns. In contrast, models like CONCH v1.5 exhibit more context-dependent preferences: they concentrate on well-differentiated glandular architectures in IDC and PRAD, but shift towards stromal, haemorrhagic, or necrotic regions in ccRCC and COAD, suggesting that AdaFusion adapts model usage according to tissue context and disease subtype. Furthermore, the results indicate a possible link between model architecture and preferred morphology; for example, Prov-GigaPath, equipped with a hierarchical slide-level encoder, often emphasises broad architectural features and interface zones between glandular and stromal tissues, particularly in PAAD and COAD.

\end{sloppypar}
\end{document}